\documentclass[sigconf]{acmart}

\AtBeginDocument{%
  }

\copyrightyear{2024}
\acmYear{2024}
\setcopyright{rightsretained}
\acmConference[KDD '24]{Proceedings of the 30th ACM SIGKDD Conference on Knowledge Discovery and Data Mining}{August 25--29, 2024}{Barcelona, Spain} \acmBooktitle{Proceedings of the 30th ACM SIGKDD Conference on Knowledge Discovery and Data Mining (KDD '24), August 25--29, 2024, Barcelona, Spain}\acmDOI{10.1145/3637528.3671581}
\acmISBN{979-8-4007-0490-1/24/08}

\makeatother

\usepackage{caption} 
\usepackage[lined,ruled,vlined, linesnumbered]{algorithm2e}
\usepackage{algpseudocode}
\SetArgSty{textnormal}
\usepackage{nicematrix}
\usepackage{multirow}
\usepackage{subfigure}
\newcommand{\update}{\textit{MemoryUpdate}}
\newcommand{\retrieval}{\textit{MemoryRetrieval}}
\newcommand{\magenta}[1]{{\color{magenta}#1}}

\begin{document}

\title{Class-incremental Learning for Time Series: \\Benchmark and Evaluation}


\author{Zhongzheng Qiao}
\email{qiao0020@e.ntu.edu.sg}
\affiliation{%
  \institution{IGP-ERI@N, NTU \\I$^2$R, A*STAR \\ CNRS@CREATE}
  \country{Singapore}
}

\author{Quang Pham}
\email{pham\_hong\_quang@i2r.a-star.edu.sg}
\author{Zhen Cao}
\email{Cao\_Zhen@i2r.a-star.edu.sg}
\affiliation{%
  \institution{I$^2$R, A*STAR}
  \country{Singapore}
}

\author{Hoang H. Le}
\email{hoangle7910@gmail.com}
\affiliation{%
 \institution{Ho Chi Minh University of Science, Vietnam National University}
 \city{Ho Chi Minh City}
 \country{Vietnam}}

\author{P. N. Suganthan}
\email{p.n.suganthan@qu.edu.qa}
\affiliation{%
  \institution{Qatar University}
  \city{Doha}
  \country{Qatar}}

\author{Xudong Jiang}
\email{exdjiang@ntu.edu.sg}
\affiliation{%
  \institution{School of Electrical and Electronic Engineering, NTU}
  \country{Singapore}}

\author{Savitha Ramasamy}
\email{ramasamysa@i2r.a-star.edu.sg}
\affiliation{%
  \institution{I$^2$R, A*STAR \\ CNRS@CREATE}
  \country{Singapore}
}


\begin{abstract}
  Real-world environments are inherently non-stationary, frequently introducing new classes over time. This is especially common in time series classification, such as the emergence of new disease classification in healthcare or the addition of new activities in human activity recognition. In such cases, a learning system is required to assimilate novel classes effectively while avoiding catastrophic forgetting of the old ones, which gives rise to the Class-incremental Learning (CIL) problem. However, despite the encouraging progress in the image and language domains, CIL for time series data remains relatively understudied. Existing studies suffer from inconsistent experimental designs, necessitating a comprehensive evaluation and benchmarking of methods across a wide range of datasets. To this end, we first present an overview of the Time Series Class-incremental Learning (TSCIL) problem, highlight its unique challenges, and cover the advanced methodologies. Further, based on standardized settings, we develop a unified experimental framework that supports the rapid development of new algorithms, easy integration of new datasets, and standardization of the evaluation process. Using this framework, we conduct a comprehensive evaluation of various generic and time-series-specific CIL methods in both standard and privacy-sensitive scenarios. Our extensive experiments not only provide a standard baseline to support future research but also shed light on the impact of various design factors such as normalization layers or memory budget thresholds. Codes are available at \magenta{\url{https://github.com/zqiao11/TSCIL}}.

\end{abstract}



\maketitle

\section{Introduction} \label{sec:intro}

Time series (TS) data play a pivotal role in various domains such as acoustics, healthcare, and manufacturing~\cite{qiu2024tfb}. Typical deep learning approaches for time series classification ~\cite{ismail2019deep} are trained on a static offline dataset, collected prior to training and under the assumption that the data are independent and identically distributed (i.i.d.). However, real-world applications often challenge this i.i.d. assumption, as practical systems usually operate in dynamic environments with non-stationary data streams where the underlying data distributions keep evolving. For instance, a TS-classification model for human activity recognition or gesture recognition should be capable of adapting to newly introduced classes~\cite{chauhan2020contauth, mahmoud2022multi}. In such scenarios, the challenge of developing an adaptive learner lies not only in seamlessly assimilating new concepts from incoming data but also in simultaneously preserving and accumulating knowledge of all encountered classes.

The primary challenge in this endeavor stems from the well-known \emph{stability-plasticity} dilemma~\cite{grossberg2012studies}, where the model must be \emph{stable} enough to remember its past knowledge, while being \emph{plastic} to accommodate new information. However, current findings~\cite{li2016learning, kirkpatrick2017overcoming} suggest that neural networks are too plastic as they cannot retain old knowledge while learning the newer ones, which is referred to as the \emph{catastrophic forgetting} phenomenon~\cite{mccloskey1989catastrophic}. Thus, developing efficient methodologies to achieve a good trade-off between facilitating learning new skills and alleviating catastrophic forgetting has played a central role in the development of continual learning (CL). Extensive efforts have been devoted to exploring various continual learning scenarios \cite{li2023crnet}, and Class-incremental Learning (CIL)~\cite{10083158, rebuffi2017icarl} emerges as the most prominent and challenging one. Nevertheless, the majority of such studies only explore image~\cite{masana2022class} or language~\cite{ke2022continual} applications. On the other hand, time series, despite its ubiquity and continuous nature, remains under investigated by the community. Existing studies suffer from inconsistency in various aspects of experimental setups, including datasets ~\cite{ehret2021continual, kiyasseh2021clinical}, normalization ~\cite{chauhan2020contauth, 10094960}, and learning protocol ~\cite{sun2023few, mahmoud2022multi}, etc. 

\begin{figure}[t]
	\begin{center}
		\includegraphics[width=.99\columnwidth]{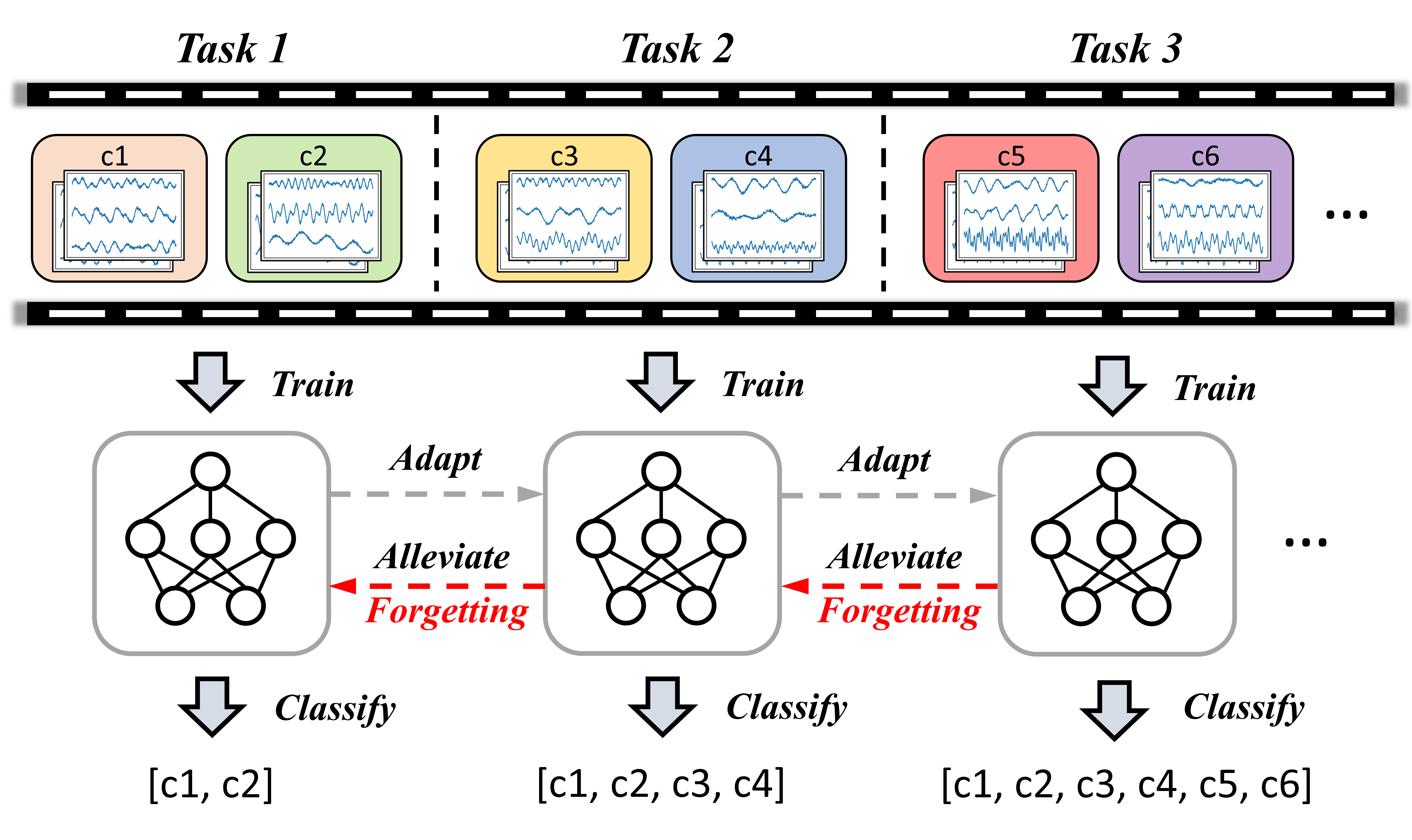}	
	\end{center}
	\caption{Schematic of Time Series Class-incremental Learning (TSCIL) process on a dynamic task sequence. Each task introduces new classes (c1 to c6), separated by clear task boundaries. The model undergoes sequential training on the tasks. After training on each task, the model needs to recognize all classes encountered thus far without catastrophic forgetting. The previously learned parameters are adapted for next task's learning.} \label{figure:cil-highlevel}
\end{figure}

To bridge this gap, this paper serves as a pioneering effort focusing exclusively on Class-incremental Learning for time series data (TSCIL). We first provide an overview of TSCIL, including problem definition, specific challenges and related works. We focus on investigating the unique characteristics of TS data such as data privacy and intra-class variations and their impact on CIL. The key contribution is the development and open-sourcing of a benchmark to facilitate a standardized evaluation of both generic and TS-specific CIL methods across various real-world datasets. This framework represents a useful resource for the research community, offering an adaptable code base for easy integration of new datasets, algorithms, and tailored learning setups, thereby empowering researchers to further develop the field of TSCIL.

Our experiments commence with the standard academic setup, evaluating both generic and TS-specific CIL methods based on regularization and experience replay \cite{de2021survey, wang2023comprehensive}. We further investigate the influence of different factors to the CIL performance, including normalization, memory budget and classifier type. Besides the standard setting, we also consider two application-specific scenarios particularly pertinent to TS modality. First, we investigate the privacy-sensitive environments, where the TS data are closely tied to individual users and unallowable to store the historical samples of previous tasks. Thus, we explore the generative replay strategy \cite{shin2017} and investigate its performance in this challenging setting. Secondly, we consider the impact of intra-class variations on TSCIL. In most datasets, time series are collected from various subjects or sources, each exhibits a distinctive input domain. Therefore, we investigate how one can incorporate such subjective information to further improve the TSCIL results.

In summary, our contributions are threefold: (1) we present a systematic overview of TSCIL, including problem definition, challenges, and existing methods. (2) We introduce a unified evaluation framework, complete with public datasets, standard protocols, and a range of methods, to facilitate further research in the field. (3) We conduct a holistic comparison of state-of-the-art CIL methodologies in both standard academic setups and application-specific scenarios, shedding light on the promises and limitations of existing methods in the context of time series data.

\section{Problem definition} \label{sec:problem_def}

\textbf{Class-incremental Learning (CIL)} involves an agent continuously learning new classes from a dynamic data stream. Following the standard academic setting \cite{kirkpatrick2017overcoming, van2019three, van2022three},  CIL represents the data stream as a sequence of tasks $\{\mathcal{T}^1, ..., \mathcal{T}^T\}$, where tasks sequentially emerge at distinct steps. Task at step $t$ is defined as $\mathcal{T}^t=\{\mathcal{D}^t, \mathcal{Y}^t\}$, characterized by a label space $\mathcal{Y}^t$ and training data $\mathcal{D}^t=\{(\mathbf{x}_i^t, y_i^t)\ | y_i^t \in \mathcal{Y}^t\}_{i=1}^{N_t}$, where $N_t$ is the number of samples. We assume that each task has the same number of disjoint classes, i.e., $|\mathcal{Y}^i| = |\mathcal{Y}^j|$ and $\mathcal{Y}^i \cap \mathcal{Y}^j = \varnothing $ for $i \neq j$. We focus on this setup with non-overlapping classes since the reappearing of old classes reduces the challenge of retaining past knowledge \cite{zhou2023deep}.

Given the task sequence, a model $f(\mathbf{x};\theta)$ is trained on all the tasks in an incremental manner. Formally, at task $\mathcal{T}^t$, we denote the model of interest by $f_{\theta_t}$, which is parameterized by $\theta_t$. The optimized parameters after learning the $t-$task is defined as $\theta^*_t$. At task $\mathcal{T}^t$, the model with parameter  $\theta^*_{t-1}$ is adapted to the new task and presented only with $\mathcal{D}^t$ for training, without access to either past or future training datasets. A memory buffer $\mathcal{M}$ with fixed budget $M$ can be optionally used, which stores a collection of historical samples for future replay (see Appendix \ref{app: alog} for details). The learning objective is to enable the model to learn the new task $\mathcal{T}^t$ effectively, while also retaining knowledge from the previous tasks \{$\mathcal{T}^1$, \dots,$\mathcal{T}^{t-1}$\}. Denoting the classification loss by $\mathcal{L}_{c}$, the ultimate learning objective to learn the entire task sequence is formulated as:

\begin{equation}\label{eq:cilobj}
    \theta^*_T = \underset{\theta}{\operatorname{argmin}} \sum_{t=1}^T \mathbb{E}_{(\mathbf{x}, y)\sim\mathcal{D}_{t}} [\mathcal{L}_{c}(f(\mathbf{x}; \theta), y)]
\end{equation}

We adapt this standard CIL setup to time series data, thereby defining the \textbf{Time Series Class-incremental Learning} (TSCIL) problem. In this setup (see Figure \ref{figure:cil-highlevel}), each sample is a time series $\mathbf{x} \in \mathbb{R}^{C\times L}$, where $C$ indicates the number of channels/variables, and $L$ denotes the length of the series. TSCIL not only inherits the constraints of standard CIL but also has its own challenges, which we highlight in the following. 

 \begin{itemize}
     \item \textbf{Normalization}: In image-based CIL, it is common to normalize images using the statistics computed from ImageNet~\cite{russakovsky2015imagenet}, scaling the pixel density to the range of $[0,1]$. However, such approaches are not directly applicable to time series due to the lack of large scale datasets encompassing many patterns. Given that data normalization is often overlooked in time series data~\cite{chauhan2020contauth, kwon2021exploring, yin2023continual}, we present a practical solution to this issue in Section \ref{subsec:protocol}.

      \item \textbf{Data Privacy}: Applications involving TS data are often associated with the need to preserve the privacy of data \cite{dwivedi2019decentralized}. This necessitates methods to avoid retaining the original user data to safeguarding privacy. Using synthetic samples has shown as a viable solution to preserve user privacy \cite{shin2017, yin2020dreaming}, which is particularly evaluated in Section \ref{subsec:exp_gr}.
     
     \item \textbf{Intra-class Variation}: Time series often exhibit greater intra-class variations than images \cite{sun2023few}. This is primarily because real-world time series are collected from various sources or subjects, each having its own characteristics~\cite{armstrong2022continual, schiemer2023online}. This phenomenon results in a complex interaction in the continual learning, where not only new classes are introduced over time, but also a class might compose several patterns. We investigate this issue in Section \ref{subsec:exp_sub}.
 \end{itemize}

 After finishing each task, the model is evaluated on the test sets of all previously learned tasks. The model needs to classify all the classes from $\mathcal{Y}^{1:t}= \mathcal{Y}^{1} \cup \dots \mathcal{Y}^{t}$, without being provided with a task identifier. The performance of the model is evaluated with metrics introduced in Section~\ref{subsec:evaluation_metrics}.

\section{Related Works}

 In existing TSCIL literature, a prevalent topic is the application of established generic CIL methods in time series scenarios. An online user authorization framework based on EWC \cite{kirkpatrick2017overcoming} and iCaRL \cite{rebuffi2017icarl} is proposed in \cite{chauhan2020contauth}, continuously recognizing new users based on biomedical TS signals. \cite{kwon2021exploring} applies classic regularization-based and replay-based methods on temporal sequences from mobile and embedded sensing applications. \cite{ehret2021continual} uses a recurrent neural network (RNN) to evaluate a variety of generic CIL methods on simple TS datasets, such as Stroke-MNIST \cite{gulcehre2017memory} and AudioSet \cite{gemmeke2017audio}. The results of these works showcase the effectiveness of using generic CIL methods to mitigate catastrophic forgetting on TS data.

Beyond adapting existing methods from image domain, innovative CIL algorithms for temporal data have also been proposed. \cite{duncker2020organizing} and \cite{yin2023continual} focus on RNN architectures and propose specific regularization-based CIL algorithms. DT$^2$W \cite{10094960} proposes a novel knowledge distillation (KD) \cite{hinton2015distilling} strategy based on soft-DTW \cite{cuturi2017soft} to mitigate stability-plasticity dilemma. 

A multitude of approaches are structured around experience replay (ER) ~\cite{rolnick2019experience, chaudhry2019continual}. CLOPS \cite{kiyasseh2021clinical} is a ER-based method for cardiac arrhythmia diagnosis, which includes an importance-based storage strategy and an uncertainty-based retrieval policy for memory buffer management. For efficient audio classification, \cite{kwon2021fasticarl} introduces a fast variant of iCaRL by replacing herding selection with KNN and utilizes quantization to compress memory samples. Using a frozen feature extractor, MAPIC \cite{sun2023few} combines a prototype enhancement module with a distance-based classifier for few-shot CIL on medical data.  

In the realm of generative replay, \cite{wang2019continual} continuously trains an autoencoder with Gaussian Mixture Models (GMM) to generate pseudo audio spectrogram data for incremental sound classification. \cite{gupta2021continual} employs separate independent generators for each task, accommodating variable input dimensions in different tasks. \cite{shevchyk2022privacy} trains separate WaveGAN \cite{donahue2018adversarial} models for different respiratory sound class and conducts a privacy evaluation on synthetic samples. Methods utilizing feature replay or prototypes are also explored. Using a fixed feature extractor, \cite{li2023few} and \cite{li2023fewtmm} update the classifier with prototypes for few-shot class-incremental audio classification.

Lastly, architecture-based techniques are also investigated. Inspired by ExpertGate \cite{aljundi2017expert}, GIM \cite{cossu2020continual} adopts a cascaded model structure which trains a task-specific RNN module for each new task. Along with RNN, a gating autoencoder is trained for each task to select the corresponding module during prediction. Moreover, \cite{sodhani2020toward} propose an expandable framework unifying GEM \cite{lopez2017gradient} and Net2Net \cite{chen2016net2net} for RNNs.

Despite the efforts made within the field, a lack of a comprehensive evaluation and comparison across various time series datasets is observed. Moreover, TSCIL suffers from inconsistencies in many crucial aspects, including datasets, learning protocols, evaluation schemes and backbones, etc. Some problematic practices related to data normalization and hyperparameter tuning even violate the fundamental principles of CIL. To address these issues, we develop a standard TSCIL framework to systematically and fairly evaluate different CIL methods on TS data.

\section{Developed Evaluation Framework}\label{sec:framework}

\subsection{Benchmark Datasets} \label{subsec:datasets}

Our TSCIL benchmarks are established with open-sourced real-world time series datasets. Based on these,  our toolkit provides a clear way to customize the CIL settings, including the number of classes per task, or the amount of training samples per class. Nevertheless, we follow a common setting in CIL studies to report the results of the balanced training setting in this paper where the amount of training samples for each class is approximately equal. We emphasize the importance of this presumption for 2 reasons. Firstly, it aligns with most standard benchmarks in conventional CIL research in vision domain \cite{van2022three} and facilitates the use of standard evaluation metrics that could be biased if classes were unbalanced. Secondly, the amount of training samples directly affects the difficulty of learning each class. Such influence can affect the performance beyond the CIL algorithm itself, therefore it is beyond the scope of this paper. 

Based on such considerations, datasets are selected from two TS-related applications: Human Activity Recognition (HAR) and Gesture Recognition. In general, a group of subjects/volunteers are asked to perform various activities or gestures for a fixed time duration. Such datasets are suitable for CIL, since there are sufficient balanced classes for task split. Some works have utilized HAR datasets for CIL \cite{jha2020continual, jha2021continual, schiemer2023online}, but they adopt the pre-processed vectors as input samples. Instead, we directly use raw time series as inputs, focusing exclusively on the TS modality. In our configuration, TS samples of each dataset exhibit a consistent shape, i.e. the sequence length and number of variables remain the same. Table \ref{tbl:datasets} shows an overview of the employed datasets.

\begin{table}[!t]
    \centering
    \caption{Overview of the benchmark datasets. The last column indicates the number of tasks in the \textit{experiment} stream.}
    \resizebox{\columnwidth}{!}{
    \begin{NiceTabular}{l|ccccc}
        \toprule
      Dataset & Shape $(C \times L$) & Train Size & Test Size & \# Classes & \# Exp Tasks  \\ \midrule
        UCI-HAR & $9\times128$ & 7352 & 2947 & 6 & 3 \\ 
        
        UWave & $3\times315$ & 896 & 3582 & 8 & 4\\
        
        DSA & $45\times125$ & 6840 & 2280 & 18 & 6\\ 
        
        GRABMyo & $28\times128$ & 36120 & 12040 & 16 & 5\\ 
        
        WISDM & $3\times200$ & 18184 & 6062 & 18 & 6\\ 
        \bottomrule
    \end{NiceTabular}}
    \label{tbl:datasets}
\end{table}

\textit{1) UCI-HAR} \cite{Ismi2016KmeansCB} contains temporal sequences of the inertial sensors of smartphones when 6 different daily activities are performed. Data are collected in 50Hz, from 30 volunteers in different ages. Sequences are directly used as inputs, which consist of 9 channels with a temporal span of 128 timesteps.

\textit{2) UWave} \cite{4912759} includes over 4000 samples collected from 8 subjects while generating 8 simple gesture patterns. We utilize the records from the three axes of the accelerometers so that each input sample is a 3-dimensional time series with 315 timesteps.

\textit{3) DSA} \cite{altun2010human} collects motion sensor segments of 19 daily sports activities carried out by 8 volunteers. Each segment, serving as a sample, is recorded across 45 distinct channels with 125 time steps. To make classes be split equally, we choose to utilize 18 classes from this dataset for experiment.

\textit{4) GRABMyo} \cite{pradhan2022multi} is a large-scaled Surface Electromyography (sEMG) database for hand-gesture recognition. It captures signals during the execution of 16 distinct gestures performed by 43 participants over three separate sessions. All recordings are 5 seconds in duration, collected from 28 channels, and sampled at 2048 Hz. We select one session's data across all subjects for experiment. We first downsample the signal to 256 Hz, followed by the application of a non-overlapping sliding window operation to cut the signal into different samples. Each window of length 0.5 second containing 128 time steps is used as an input sample. We aggregate all of the windows from each subject and perform train-test split with a 3:1 ratio, ensuring that both training and test data are from all the subjects. This avoids introducing distribution shifts caused by subjects between train and test data, appropriate for our focus on CIL. The Offline  results in Table \ref{tbl:main_results} indicate that our processed samples contain sufficient information for class differentiation.

\textit{5) WISDM} \cite{weiss2019wisdm} is a sensor-based HAR dataset encompassing 18 activities and involving 51 subjects. Following \cite{yin2023continual}, we utilize the phone accelerator modality and extract samples by applying a non-overlapping sliding window with a window size of 200. Each sample comprises a 10-second time series with a frequency of 20 Hz. Similar to the practice for GrabMyo, the dataset is divided into training and test sets with a 3:1 ratio, making both sets include data from all the subjects.

\begin{figure*}[!t]
	\begin{center}
		\subfigure[BatchNorm]
		{	\includegraphics[width=1.9\columnwidth]{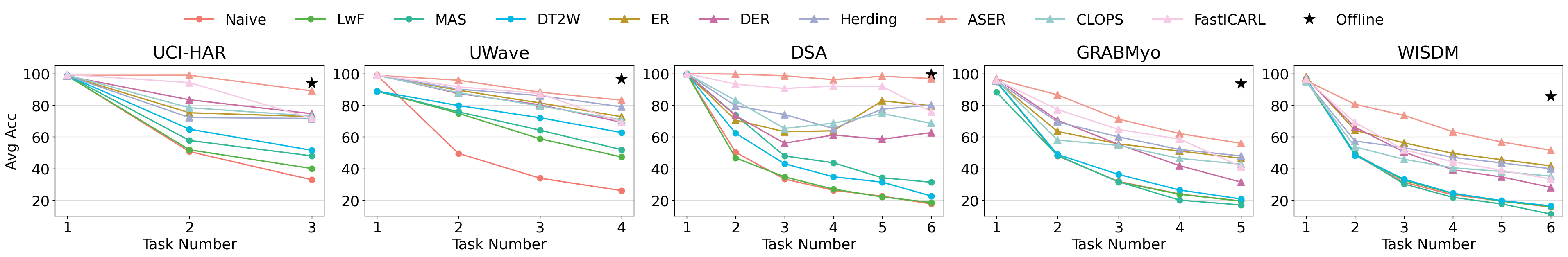}
	   \label{figure:acc_evol_BN}	
  }

		\subfigure[LayerNorm]
		{	\includegraphics[width=1.9\columnwidth]{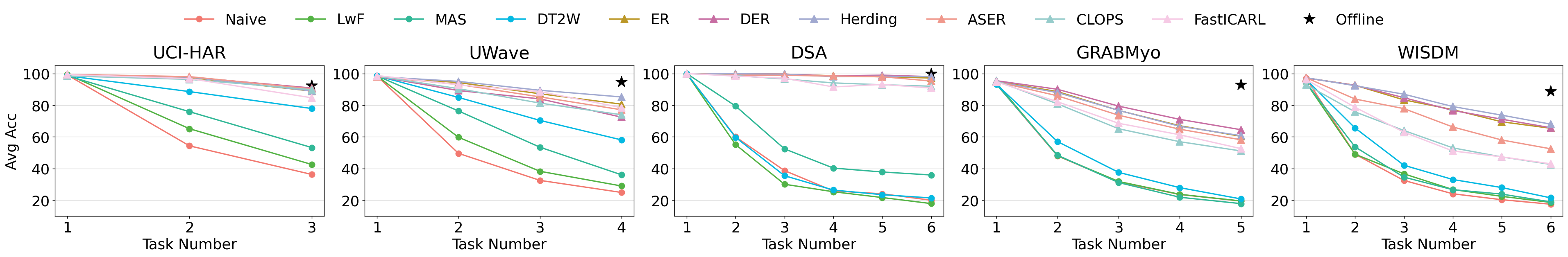}
            \label{figure:acc_evol_LN}
		}

	\end{center}
 	\vspace{-5mm}
	\caption{Evolution of Average Accuracy ($\mathcal{A}_i$) when using (a) BatchNorm or (b) LayerNorm for normalization. Methods utilizing memory buffer are marked with triangles. Since Offline represents joint training on the entire task sequence, its result shows as a single point instead of a curve.
	} \label{figure:acc_evol}
\end{figure*}

\subsection{Learning Protocols} \label{subsec:protocol}

\subsubsection{Task Split} Following the standard CIL definition, we need to split the dataset into $T$ tasks, ensuring that each task contains mutually exclusive classes. Similar to the procedure in \cite{rebuffi2017icarl}, we shuffle the class order before splitting. That enables us to assess the robustness of CIL methods against class order. After that, we split all the classes equally into each task. Similar to Split-MNIST and Split-CIFAR10 \cite{van2019three}, we allocate 2 distinct classes to each task within this work.

 \subsubsection{Data Normalization} Normalization of input data is crucial for model's training. Many TSCIL studies apply Z-score normalization before task split and use the statistics calculated on the entire dataset for normalization~\cite{chauhan2020contauth, kwon2021exploring, yin2023continual}. This practice violates the fundamental principle of CL as the full dataset is not accessible prior to training. To address this issue, we apply instance-wise normalization by inserting an input normalization layer before the first layer of the model. It can be LayerNorm (LN) \cite{ba2016layer} or InstanceNorm (IN) \cite{ulyanov2016instance}, \textit{without} incorporating a learnable affine transformation or bias. This ensures inputs are normalized to have a mean of 0 and a standard deviation of 1 along specific dimensions. The choice of input normalization layer can be decided based on the performance on the validation tasks. Except for WISDM, where no normalization is applied, we apply IN for UWave, while LN is employed for the remaining datasets.

  \subsubsection{Hyperparameter Tuning} The selection of hyperparameters is a challenging issue in the realm of CL, typically following two protocols. The first \cite{kirkpatrick2017overcoming, shin2017} involves dividing each task into train, validation, and test sets, then performing  grid search. The best parameters are chosen based on validation set performance across all tasks. However, this method, requiring visits to the entire task stream and necessitates a strong assumption of the relatedness between the prior validation data and future tasks. Another protocol~\cite{chaudhry2018efficient} divides tasks into a 'validation' stream for cross-validation and hyperparameter tuning, and an 'experiment' stream for training and evaluation. We use the first protocol for UCI-HAR and UWave (having only 3 and 4 tasks, respectively) and the second for datasets with more tasks, setting the validation stream task count to 3. We highlight that both protocols are common standard practices, each with its own advantages and limitations. We offer both options in our toolkit, allowing users to choose based on their needs.


\begin{table}[!b]
    \centering
    \caption{Summary of the implemented CIL algorithms.}
    \resizebox{\columnwidth}{!}{
    \begin{NiceTabular}{l|ccc}
        \toprule
        Algorithm & Application & Category & Characteristics  \\ \midrule
        
        LwF \cite{li2016learning} & Visual & Regularization & KD on logits \\ 
        
        MAS \cite{aljundi2018memory} & Visual & Regularization & Parameter regularization\\
        
        DT$^2$W \cite{10094960} & Time Series & Regularization & KD on feature maps\\ 

        \hline
        
        ER \cite{rolnick2019experience} & Visual & Experience &  Replay baseline\\ 

        Herding \cite{rebuffi2017icarl} & Visual & Experience &  Memory update \\

        DER \cite{buzzega2020dark} & Visual & Experience&  Replay with logits\\
        
        ASER \cite{shim2021online} & Visual & Experience & Memory retrieval\\ 

        CLOPS \cite{kiyasseh2021clinical} & Time Series & Experience &  Update \& Retrieval \\

        FastICARL \cite{kwon2021fasticarl} & Time Series & Experience &  Memory Update \\
        
        \hline
        
        GR \cite{shin2017} & Visual & Generative & Auxiliary generator\\ 
        
        
        \bottomrule
    \end{NiceTabular}}
    \label{tbl:comp_methods}
\end{table}

\subsection{Selected Methods}

 We firstly select 9 representative methods based on regularization and experience replay techniques for comparison. These methods include generic methods proposed in image domain as well as specific algorithms for TS data. Among the regularization-based methods, we choose \textbf{LwF} \cite{li2016learning}, \textbf{MAS} \cite{aljundi2018memory} and \textbf{DT$^2$W} \cite{10094960}. For experience replay, \textbf{ER} \cite{rolnick2019experience}, \textbf{DER} \cite{buzzega2020dark}, \textbf{Herding} \cite{rebuffi2017icarl}, \textbf{ASER} \cite{shim2021online}, \textbf{CLOPS} \cite{kiyasseh2021clinical} and \textbf{FastICARL} \cite{kwon2021fasticarl} are included. To investigate the scenario with data privacy concerns, we further incorporate a generative-replay-based method: \textbf{GR} \cite{shin2017}. This method avoids saving raw samples, and its experiment results are discussed in Section \ref{subsec:exp_gr}. 
Lastly, we report the results of two straightforward baselines: \textbf{Naive} and \textbf{Offline}. The former gives the lower bound of performance, as it finetunes the model on tasks sequentially without using any CIL technique. The later one serves as the ideal upper bound since it undergoes joint training with all samples from the entire data stream. A summary of selected CIL methods is presented in Table \ref{tbl:comp_methods}. Additional details are included in Appendix \ref{app: alog}.

\subsection{Evaluation Metrics}
\label{subsec:evaluation_metrics}

We employ 3 standard metrics for TSCIL evaluation. Let $a_{i,j}$ represent the accuracy evaluated on the test set of learned task $j \leq i$ upon trained task $i$. (1) \textbf{Average Accuracy} after learning task $i$ is defined as $\mathcal{A}_i = \frac{1}{i} \sum_{j=1}^{i} a_{i,j}$, which is the mean of accuracies across all test sets of learned tasks and reflects the model's overall performance. (2) \textbf{Average Forgetting} \cite{chaudhry2018efficient} after leaning task $i$ is defined as $\mathcal{F}_i = \frac{1}{i-1}\sum_{j=1}^{i-1} f_{i,j}$, where $f_{i,j}=\max_{k \in \{1,...,i-1\}} (a_{k,j}) - a_{i, j}, j<i$ represents how much performance degrades on task $j$ due to learning task $i$. This metric reflects how much of the acquired knowledge the model has forgotten in a task-level. (3) \textbf{Average Learning Accuracy} \cite{riemer2018learning} is defined as $\mathcal{A}_{cur} = \frac{1}{T} \sum_{i=1}^{T} a_{i,i}$. 
This metric indicates the overall impact of using a CIL method on learning new tasks, which is reflected by the average of \textit{current task accuracy} $a_{i,i}$ across all the tasks in sequence. To reflect the final performance, the community commonly reports \textbf{Final Average Accuracy} $\mathcal{A}_T$ and \textbf{Final Average Forgetting} $\mathcal{F}_T$, which are computed across all the tasks after the final task has been learned.

\subsection{Model Architecture}  \label{subsec:model}
For the experiments in this paper, we employ a \textbf{1D-CNN} backbone similar to \cite{ragab2023adatime} as the feature extractor. It consists of four convolutional blocks, with each block comprising a 1D-Convolutional layer, a BatchNorm (BN) layer, a MaxPooling layer and a Dropout Layer. Unless otherwise stated, we utilize a single-head classifier with softmax activation across all the algorithms. We specifically investigate the impact of different types of classifier in the ablation study. For methods using memory buffer, we set the buffer size to 5\% of the training size in the experiment task stream.
Furthermore, normalization layers also play a vital role in CIL problems. Although most literature incorporate BN layers into their models, it has been empirically shown that BN layers suffer from the biased issue in CIL scenarios ~\cite{pham2021continual}. We further investigate this issue in the realm of TSCIL by comparing the results of using BN and LN. 
For the generator of GR, we utilize a TimeVAE \cite{desai2021timevae}, with both the encoder and decoder designed with four layers of Conv1D and ConvTranspose1d, respectively.

\subsection{Implementation Details} \label{subsec:implementation}

All the experiments are run 5 times with different class orders and random seeds. For each run, we tune its particular best hyperparameters as the protocols described above. Similar to \cite{van2022three}, all models are trained for 100 epochs per task using an Adam optimizer with a learning rate of 0.001 and a batch size of 64. The learning rate scheduler is configured as a hyperparameter for tuning. To alleviate overfitting on training data, early stopping is used during training. For a fair comparison, we choose not to tune the architecture-related parameters for different methods. Instead, we employ a fixed and consistent model architecture. Further details on implementation of the framework can be found in Appendix \ref{app: imple details}. We highlight that our framework is extensible. Users can follow the instructions in our code page to incorporate new datasets, algorithms, and custom experimental setups.

\begin{table*}[!ht]
    \centering
    \caption{Evaluation metrics of regularization-based and ER-based methods on our 5 TSCIL benchmarks when using (a) BatchNorm or (b) LayerNorm for normalization. Metrics introduced in Section \ref{subsec:evaluation_metrics} are reported, which are $\mathcal{A}_T$($\uparrow$), $\mathcal{F}_T$($\downarrow$) and $\mathcal{A}_{cur}$($\uparrow$). For each metric, its mean and confidence interval on 5 runs are reported.} 
    \vspace{-3.5mm}
    \subtable[BatchNorm]{
    \resizebox{\textwidth}{!}{
    \begin{NiceTabular}{c|c|cc|ccc|cccccc}
        \toprule
                     Dataset & Metric & Naive & Offline & LwF & MAS & DT$^2$W & ER & DER & Herding & ASER & CLOPS & FastICARL \\ \midrule
        \multirow{3}{*}{UCI-HAR} & $\mathcal{A}_T$ & 32.9{\tiny $\pm$1.1} & 93.9{\tiny $\pm$0.6} & 40.0{\tiny $\pm$7.4} & 48.0{\tiny $\pm$11.4} & \textbf{51.5}{\tiny $\pm$14.1} & 72.8{\tiny $\pm$19.7} & 74.6{\tiny $\pm$3.5} & 71.6{\tiny $\pm$23.9} & \textbf{89.1}{\tiny $\pm$6.1} & 73.2{\tiny $\pm$10.5} & 71.7{\tiny $\pm$19.1}  \\
                       ~ & $\mathcal{F}_T$ & 97.8{\tiny $\pm$3.3} & N.A & 83.0{\tiny $\pm$14.1} & 71.6{\tiny $\pm$11.8} & \textbf{62.0}{\tiny $\pm$16.3} & 22.1{\tiny $\pm$37.2} & 16.6{\tiny $\pm$18.8} & 25.1{\tiny $\pm$44.8} & \textbf{13.2}{\tiny $\pm$11.4} & 18.4{\tiny $\pm$25.6} & 32.4{\tiny $\pm$34.0} \\
                       ~ & $\mathcal{A}_{cur}$ & 98.1{\tiny $\pm$3.3} & N.A & \textbf{95.4}{\tiny $\pm$5.6} & 93.2{\tiny $\pm$9.5} & 92.8{\tiny $\pm$6.2} & 87.2{\tiny $\pm$10.4} & 85.6{\tiny $\pm$11.7} & 88.3{\tiny $\pm$9.0} & \textbf{97.8}{\tiny $\pm$1.9} & 85.2{\tiny $\pm$7.7} & 93.3{\tiny $\pm$5.7} \\ \midrule

        \multirow{3}{*}{UWave} & $\mathcal{A}_T$ & 26.0{\tiny $\pm$3.0} & 96.6{\tiny $\pm$0.7} & 47.3{\tiny $\pm$11.1} & 51.9{\tiny $\pm$11.1} & \textbf{62.7}{\tiny $\pm$11.6} & 72.7{\tiny $\pm$2.7} & 69.2{\tiny $\pm$4.5} & 79.0{\tiny $\pm$1.1} & \textbf{83.2}{\tiny $\pm$3.5} & 70.8{\tiny $\pm$3.0} & 69.3{\tiny $\pm$2.4} \\
                             ~ & $\mathcal{F}_T$ & 97.3{\tiny $\pm$3.7} & N.A & 51.7{\tiny $\pm$19.2} & 45.7{\tiny $\pm$15.1} & \textbf{32.9}{\tiny $\pm$12.0} & 33.6{\tiny $\pm$3.4} & 37.8{\tiny $\pm$6.9} & 24.7{\tiny $\pm$2.4} & \textbf{20.3}{\tiny $\pm$4.5} & 36.1{\tiny $\pm$3.7} & 37.8{\tiny $\pm$4.2} \\
                             ~ & $\mathcal{A}_{cur}$ & 99.0{\tiny $\pm$0.6} & N.A & \textbf{86.1}{\tiny $\pm$14.1} & 78.3{\tiny $\pm$16.2} & 83.6{\tiny $\pm$17.0} & 97.9{\tiny $\pm$0.5} & 97.6{\tiny $\pm$1.0} & 97.5{\tiny $\pm$1.4} & \textbf{98.4}{\tiny $\pm$0.4} & 97.8{\tiny $\pm$0.4} & 97.6{\tiny $\pm$0.9} \\ \midrule

        \multirow{3}{*}{DSA} & $\mathcal{A}_T$ & 17.6{\tiny $\pm$2.7} & 99.5{\tiny $\pm$0.6} & 18.5{\tiny $\pm$6.4} & \textbf{31.3}{\tiny $\pm$6.4} & 22.6{\tiny $\pm$6.6} & 79.6{\tiny $\pm$15.6} & 62.7{\tiny $\pm$23.0} & 80.1{\tiny $\pm$6.0} & \textbf{96.9}{\tiny $\pm$2.2} & 68.5{\tiny $\pm$20.5} & 75.9{\tiny $\pm$21.1} \\
                           ~ & $\mathcal{F}_T$ & 98.8{\tiny $\pm$3.2} & N.A & 85.1{\tiny $\pm$22.7} & \textbf{60.0}{\tiny $\pm$11.5} & 87.5{\tiny $\pm$8.2} & 23.6{\tiny $\pm$19.2} & 21.4{\tiny $\pm$17.4} & 20.4{\tiny $\pm$6.5} & \textbf{3.7}{\tiny $\pm$2.7} & 32.1{\tiny $\pm$27.4} & 25.1{\tiny $\pm$26.4} \\
                           ~ & $\mathcal{A}_{cur}$ & 100.0{\tiny $\pm$0.0} & N.A  & 87.8{\tiny $\pm$29.0} & 81.2{\tiny $\pm$16.3} & \textbf{95.5}{\tiny $\pm$1.7} & 89.9{\tiny $\pm$9.6} & 76.9{\tiny $\pm$20.0} &90.9{\tiny $\pm$7.1} & \textbf{100.0}{\tiny $\pm$0.1} & 88.0{\tiny $\pm$9.0} & 95.3{\tiny $\pm$1.2} \\ \midrule

       \multirow{3}{*}{GRABMyo} & $\mathcal{A}_T$ & 19.4{\tiny $\pm$0.3} & 93.8{\tiny $\pm$1.0} & 19.4{\tiny $\pm$0.6} & 16.9{\tiny $\pm$2.2} & \textbf{20.7}{\tiny $\pm$5.2} & 46.5{\tiny $\pm$3.2} & 31.4{\tiny $\pm$3.7} & 47.9{\tiny $\pm$3.9} & \textbf{55.9}{\tiny $\pm$5.1} & 42.7{\tiny $\pm$4.3} & 41.3{\tiny $\pm$5.3} \\
                                ~ & $\mathcal{F}_T$ & 95.4{\tiny $\pm$1.6} & N.A & 95.0{\tiny $\pm$1.5} & \textbf{49.8}{\tiny $\pm$25.3} & 59.3{\tiny $\pm$19.0} & 35.0{\tiny $\pm$5.7} & 59.8{\tiny $\pm$8.2} & 33.1{\tiny $\pm$8.7} & 49.4{\tiny $\pm$5.0} & \textbf{31.1}{\tiny $\pm$4.9} & 47.0{\tiny $\pm$5.8} \\
                                ~ & $\mathcal{A}_{cur}$ & 95.8{\tiny $\pm$1.2} & N.A & \textbf{95.4}{\tiny $\pm$1.1} & 56.8{\tiny $\pm$20.9} & 67.2{\tiny $\pm$17.7} & 73.7{\tiny $\pm$4.7} & 79.2{\tiny $\pm$7.3} & 73.2{\tiny $\pm$6.2} & \textbf{95.4}{\tiny $\pm$1.3} & 65.1{\tiny $\pm$5.6} & 78.4{\tiny $\pm$5.9} \\ \midrule

       \multirow{3}{*}{WISDM} & $\mathcal{A}_T$ & 15.5{\tiny $\pm$1.2} & 85.7{\tiny $\pm$1.9} & 15.9{\tiny $\pm$0.7} & 11.2{\tiny $\pm$5.3} & \textbf{16.4}{\tiny $\pm$6.6} & 41.7{\tiny $\pm$5.0} & 28.1{\tiny $\pm$10.1} & 39.9{\tiny $\pm$10.6} & \textbf{51.6}{\tiny $\pm$13.7} & 35.1{\tiny $\pm$6.1} & 33.4{\tiny $\pm$4.8} \\
                                ~ & $\mathcal{F}_T$ & 95.6{\tiny $\pm$3.4} & N.A & 96.4{\tiny $\pm$2.7} & 78.0{\tiny $\pm$8.2} & \textbf{63.8}{\tiny $\pm$44.0} & \textbf{27.9}{\tiny $\pm$12.4} & 55.6{\tiny $\pm$24.1} & 28.6{\tiny $\pm$19.7} & 53.1{\tiny $\pm$14.3} & 33.7{\tiny $\pm$13.7} & 39.7{\tiny $\pm$6.8} \\
                                ~ & $\mathcal{A}_{cur}$ & 95.2{\tiny $\pm$3.1} & N.A & \textbf{96.2}{\tiny $\pm$1.8} & 76.2{\tiny $\pm$5.4} & 69.3{\tiny $\pm$38.2} & 60.7{\tiny $\pm$10.2} & 74.5{\tiny $\pm$18.0} & 61.3{\tiny $\pm$10.1} & \textbf{95.8}{\tiny $\pm$2.0} & 60.3{\tiny $\pm$10.3} & 66.2{\tiny $\pm$5.8} \\
       \bottomrule
    \end{NiceTabular}}
    \label{tbl:main_results_BN}
    }

    \hfill

    \subtable[LayerNorm]{
    \resizebox{\textwidth}{!}{
    \begin{NiceTabular}{c|c|cc|ccc|cccccc}
        \toprule
                     Dataset & Metric & Naive & Offline & LwF & MAS & DT$^2$W & ER & DER & Herding & ASER & CLOPS & FastICARL \\ \midrule
        \multirow{3}{*}{UCI-HAR} & $\mathcal{A}_T$ & 36.2{\tiny $\pm$10.9} & 92.5{\tiny $\pm$0.8} & 42.5{\tiny $\pm$14.2} & 53.2{\tiny $\pm$7.5} & \textbf{77.9}{\tiny $\pm$15} & 88.9{\tiny $\pm$2.7} & \textbf{90.9}{\tiny $\pm$1.7} & 89.0{\tiny $\pm$1.9} & 90.3{\tiny $\pm$1.9} & 89.8{\tiny $\pm$1.5} & 84.7{\tiny $\pm$4.8} \\
                                ~ & $\mathcal{F}_T$ & 92.5{\tiny $\pm$13.5} & N.A & 83.1{\tiny $\pm$19.0} & 64.8{\tiny $\pm$18.8} & \textbf{9.3}{\tiny $\pm$6.6} & 10.8{\tiny $\pm$6.8} & \textbf{8.5}{\tiny $\pm$5.7} & 10.7{\tiny $\pm$6.5} & 9.6{\tiny $\pm$5.7} & 9.2{\tiny $\pm$4.9} & 18.2{\tiny $\pm$10.7} \\
                                ~ & $\mathcal{A}_{cur}$ & 97.9\textbf{}{\tiny $\pm$3.5} & N.A & \textbf{97.9}{\tiny $\pm$3.8} & 95.8{\tiny $\pm$4.5} & 83.7{\tiny $\pm$19.7} & 96.1{\tiny $\pm$2.8} & 96.3{\tiny $\pm$2.7} & 96.1{\tiny $\pm$3.2} & \textbf{96.6}{\tiny $\pm$1.8} & 95.9{\tiny $\pm$2.6} & 96.8{\tiny $\pm$3.6} \\ \midrule

        \multirow{3}{*}{UWave} & $\mathcal{A}_T$ & 24.8{\tiny $\pm$0.1} & 94.7{\tiny $\pm$0.7} & 28.9{\tiny $\pm$4.6} & 36.0{\tiny $\pm$8.1} & \textbf{58.1}{\tiny $\pm$18.3} & 80.7{\tiny $\pm$1.4} & 72.4{\tiny $\pm$18.0} & \textbf{85.2}{\tiny $\pm$2.2} & 77.1{\tiny $\pm$7.5} & 74.0{\tiny $\pm$2.1} & 78.5{\tiny $\pm$0.8} \\
                                ~ & $\mathcal{F}_T$ & 98.4{\tiny $\pm$1.5} & N.A & 69.8{\tiny $\pm$38.6} & 71.5{\tiny $\pm$11.1} & \textbf{35.4}{\tiny $\pm$25.6} & 22.7{\tiny $\pm$1.3} & 33.5{\tiny $\pm$22.1} & \textbf{16.6}{\tiny $\pm$2.4} & 28.1{\tiny $\pm$9.1} & 32.2{\tiny $\pm$3.2} & 26.4{\tiny $\pm$0.8} \\
                                ~ & $\mathcal{A}_{cur}$ & 98.6{\tiny $\pm$1.1} & N.A & 81.2{\tiny $\pm$25.0} & \textbf{89.7}{\tiny $\pm$4.2} & 87.3{\tiny $\pm$2.6} & 97.7{\tiny $\pm$1.0} & 97.5{\tiny $\pm$1.5} & 97.5{\tiny $\pm$1.4} & \textbf{98.2}{\tiny $\pm$0.9} & 98.1{\tiny $\pm$0.7} & 98.3{\tiny $\pm$0.4} \\ \midrule

        \multirow{3}{*}{DSA} & $\mathcal{A}_T$ & 19.9{\tiny $\pm$5.1} & 99.8{\tiny $\pm$0.1} & 17.8{\tiny $\pm$4.2} & \textbf{35.9}{\tiny $\pm$5.6} & 21.4{\tiny $\pm$7.1} & 97.2{\tiny $\pm$2.2} & \textbf{98.0}{\tiny $\pm$1.0} & 98.0{\tiny $\pm$1.8} & 95.3{\tiny $\pm$4.7} & 92.0{\tiny $\pm$4.1} & 90.8{\tiny $\pm$3.0} \\
                                ~ & $\mathcal{F}_T$ & 96.1{\tiny $\pm$6.2} & N.A & 89.9{\tiny $\pm$12.7} & \textbf{53.0}{\tiny $\pm$9.1} & 94.0{\tiny $\pm$8.7} & 3.3{\tiny $\pm$2.7} & \textbf{2.2}{\tiny $\pm$1.2} & 2.3{\tiny $\pm$2.2} & 5.6{\tiny $\pm$5.7} & 9.4{\tiny $\pm$5.0} & 10.9{\tiny $\pm$3.6} \\
                                ~ & $\mathcal{A}_{cur}$ & 100.0{\tiny $\pm$0.0} & N.A & 92.5{\tiny $\pm$12.3} & 80.0{\tiny $\pm$9.5} & \textbf{99.7}{\tiny $\pm$0.6} & 99.9{\tiny $\pm$0.1} & 99.9{\tiny $\pm$0.2} & 99.9{\tiny $\pm$0.1} & 99.9{\tiny $\pm$0.2} & 99.9{\tiny $\pm$0.2} & 99.9{\tiny $\pm$0.1} \\ \midrule

        \multirow{3}{*}{GRABMyo} & $\mathcal{A}_T$ & 19.4{\tiny $\pm$0.3} & 92.9{\tiny $\pm$2.3} & 19.5{\tiny $\pm$0.2} & 17.8{\tiny $\pm$1.2} & \textbf{20.8}{\tiny $\pm$7.1} & 60.3{\tiny $\pm$1.1} & \textbf{64.5}{\tiny $\pm$3.5} & 60.7{\tiny $\pm$2.7} & 58.1{\tiny $\pm$3.9} & 51.1{\tiny $\pm$3.5} & 52.6{\tiny $\pm$4.2} \\
                                ~ & $\mathcal{F}_T$ & 94.4{\tiny $\pm$2.4} & N.A & 94.7{\tiny $\pm$2.6} & 84.2{\tiny $\pm$3.5} & \textbf{21.9}{\tiny $\pm$10.4} & 42.1{\tiny $\pm$1.6} & \textbf{36.6}{\tiny $\pm$3.4} & 41.7{\tiny $\pm$3.3} & 45.3{\tiny $\pm$4.7} & 53.6{\tiny $\pm$4.5} & 52.1{\tiny $\pm$5.6} \\
                                ~ & $\mathcal{A}_{cur}$ & 94.9{\tiny $\pm$1.7} & N.A & \textbf{95.2}{\tiny $\pm$1.9} & 85.2{\tiny $\pm$3.0} & 36.0{\tiny $\pm$11.2} & 94.0{\tiny $\pm$0.8} & 93.8{\tiny $\pm$0.8} & 94.1{\tiny $\pm$1.4} & \textbf{94.4}{\tiny $\pm$1.1} & 93.9{\tiny $\pm$1.4} & 94.2{\tiny $\pm$2.0} \\ \midrule

        \multirow{3}{*}{WISDM} & $\mathcal{A}_T$ & 17.4{\tiny $\pm$3.4} & 88.7{\tiny $\pm$1.8} & 18.5{\tiny $\pm$4.3} & 18.9{\tiny $\pm$5.0} & \textbf{21.4}{\tiny $\pm$6.6} & 65.4{\tiny $\pm$5.8} & 65.8{\tiny $\pm$6.9} & \textbf{68.0}{\tiny $\pm$7.6} & 52.4{\tiny $\pm$17.1} & 42.6{\tiny $\pm$7.7} & 42.9{\tiny $\pm$5.1} \\ 
                                ~ & $\mathcal{F}_T$ & 95.4{\tiny $\pm$5.5} & N.A & 91.9{\tiny $\pm$7.5} & 75.6{\tiny $\pm$12.2} & \textbf{41.2}{\tiny $\pm$21.5} & 34.7{\tiny $\pm$5.4} & 35.2{\tiny $\pm$7.3} & \textbf{31.9}{\tiny $\pm$7.9} & 52.4{\tiny $\pm$20.6} & 62.3{\tiny $\pm$8.2} & 57.3{\tiny $\pm$9.2} \\ 
                                ~ & $\mathcal{A}_{cur}$ & 97.0{\tiny $\pm$1.6} & N.A & \textbf{95.1}{\tiny $\pm$3.8} & 81.8{\tiny $\pm$7.7} & 49.4{\tiny $\pm$22.1} & 94.3{\tiny $\pm$1.9} & 95.1{\tiny $\pm$1.2} & 94.6{\tiny $\pm$1.2} & \textbf{96.1}{\tiny $\pm$1.2} & 94.5{\tiny $\pm$1.2} & 90.7{\tiny $\pm$6.3} \\ 

        \bottomrule
    \end{NiceTabular}}
    \label{tbl:main_results_LN}
    }

\end{table*}

\section{Experiments and Discussion} \label{sec:experiment}

\subsection{Evaluation of Regularization-based and ER-based Methods}
\label{subsec: exp_reg_er}

\subsubsection{Performance comparison using BN and LN}

We first focus on a basic scenario in which saving historical samples is permitted. As listed in Table \ref{tbl:comp_methods}, we evaluate 3 regularization-based methods and 5 methods rooted on experience replay. Simultaneously, we also investigate the influence of normalization layers by running 2 sets of experiments. One uses the default CNN backbone with BN layers, the other replaces the BN layers in the model with LN layers. The results of overall performances are shown in Table \ref{tbl:main_results_BN} for BN and Table \ref{tbl:main_results_LN} for LN. We also present the evolution of Average Accuracy $\mathcal{A}_i$ across tasks in Figure \ref{figure:acc_evol}. The evaluation results answer the following questions.

\begin{figure*}[t]
	\begin{center}
		\subfigure[BatchNorm]
		{	\includegraphics[width=1.9\columnwidth]{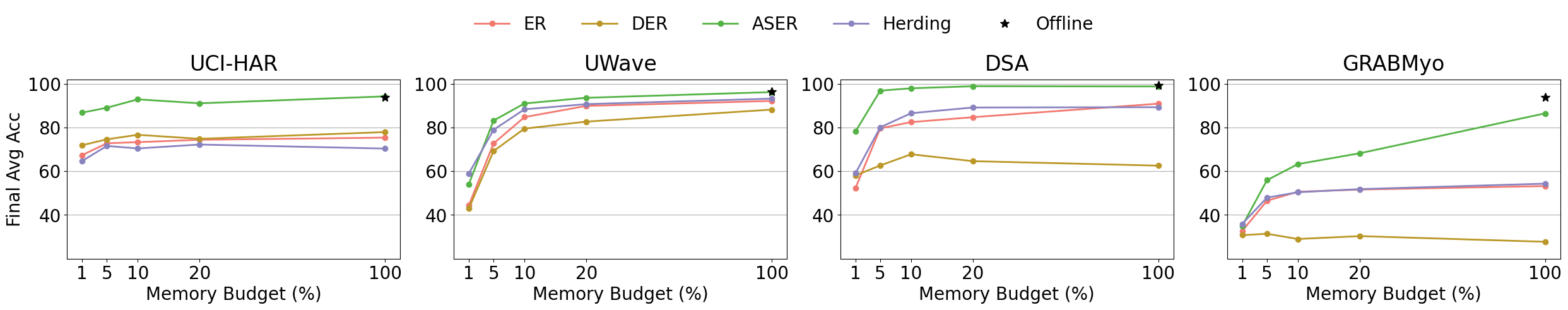}
	\label{figure:abl_mem_bn}	
  }
		\subfigure[LayerNorm]
		{	\includegraphics[width=1.9\columnwidth]{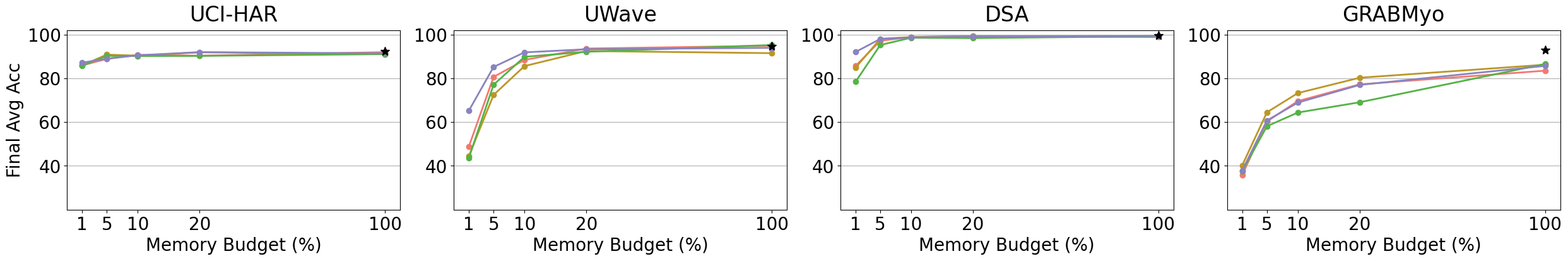}		
		\label{figure:abl_mem_ln}
  }
	\end{center}
 	\vspace{-5mm}
	\caption{Evolution of Final Average Accuracy ($\mathcal{A}_T$) when using different memory budget. The results encompass 4 ER-based methods on 4 datasets, utilizing BatchNrom (top row) or LayerNorm (bottom row) for normalization. 
	} \label{figure:abl_mem}
\label{tbl:main_results}
\end{figure*}

\textbf{Question 1:} \textbf{How do regularization vs ER perform in TSCIL?} Similar to the findings in image domain \cite{mai2022online, li2024towards}, \textit{ER-based methods stably outperform regularization-based methods without saving exemplars in TSCIL}. As anticipated, without using any CIL techniques, Naive inevitably results in catastrophic forgetting. By saving memory samples, all the ER-based methods effectively mitigate forgetting across datasets. Surprisingly, when using LN for normalization, even the basic ER method with a 5\% memory budget can sometimes achieve close results to the offline training upper bound. In contrast, regularization-based methods only show clear benefits in simpler benchmarks with fewer tasks like UCI-HAR and UWave. In these datasets, DT$^2$W consistently outperforms MAS, which in turn offers better results than LwF. However, in more challenging benchmarks, the same regularization approaches fail almost completely. Specifically, they struggle with balancing stability and plasticity, leading to either significant forgetting (LwF) or compromised learning accuracy (MAS and DT$^2$W).

\textbf{Question 2:} \textbf{How does the choice of BN and LN affects TSCIL?} While the choice between BN and LN has a marginal impact on offline training, we find that \textit{using LN appears to significantly improve the performance for most ER-based methods}. Remarkably, the impact of switching to LN is so profound that it can overshadow the choice of the algorithm itself. In some cases, merely transitioning from BN to LN within the same algorithm can boost the performance to levels almost on par with offline training. \cite{pham2021continual} attributes this phenomenon to the bias of running statistics in BN, which is cased by the imbalance of new and memory samples and results in a loss of previously acquired knowledge. In contrast, employing instance-wise normalization, such as LN, effectively circumvents this issue. However, we emphasize that the influence of the bias of BN is \textit{bidirectional}. Based on the change of learning accuracy $\mathcal{A}_{cur}$, we find that the bias of BN not only degrades the stability, but also hinders the learning of new knowledge. Additionally, the bias of BN exerts a more pronounced effect on replay with logits compared to replay with raw samples, with DER experiencing significant improvement upon substituting BN with LN. Interestingly, ASER appears as an exemption: its performance on BN is extensively better than other compared methods, yet it does not show notable benefit from using LN. We posit that this is due to the \textit{MemoryRetrieval} mechanism of ASER, which selects a balanced and representative batch of memory samples to maintain an unbiased statistics in BN layers. In some extent, the superiority of ASER on BN underscores the significance of \textit{MemoryRetrieval} within ER techniques. Contrary to ER-based methods, regularization-based methods fail to exhibit a consistent pattern across BN and LN. In conclusion, ER-based methods consistently benefit from using LN with a substantial improvement, whereas regularization-based methods need to decide the choice of BN or LN based on the dataset.

\subsubsection{Ablation study} 
\label{subsubsec: exp_ablation}

\begin{figure}[t]
    \begin{center}
        \subfigure[BatchNorm]
        {
            \includegraphics[width=.99\columnwidth]{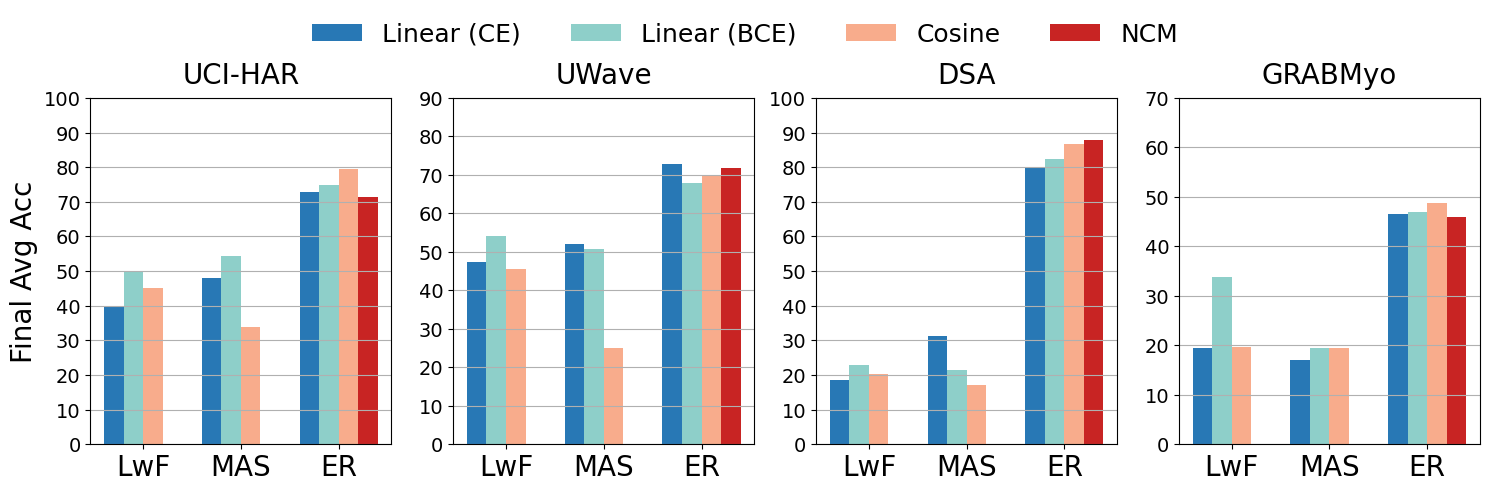}
            \label{figure:abl_cls_bn}
        }
    \end{center}
    \begin{center}
        \subfigure[LayerNorm]
        {
            \includegraphics[width=.99\columnwidth]{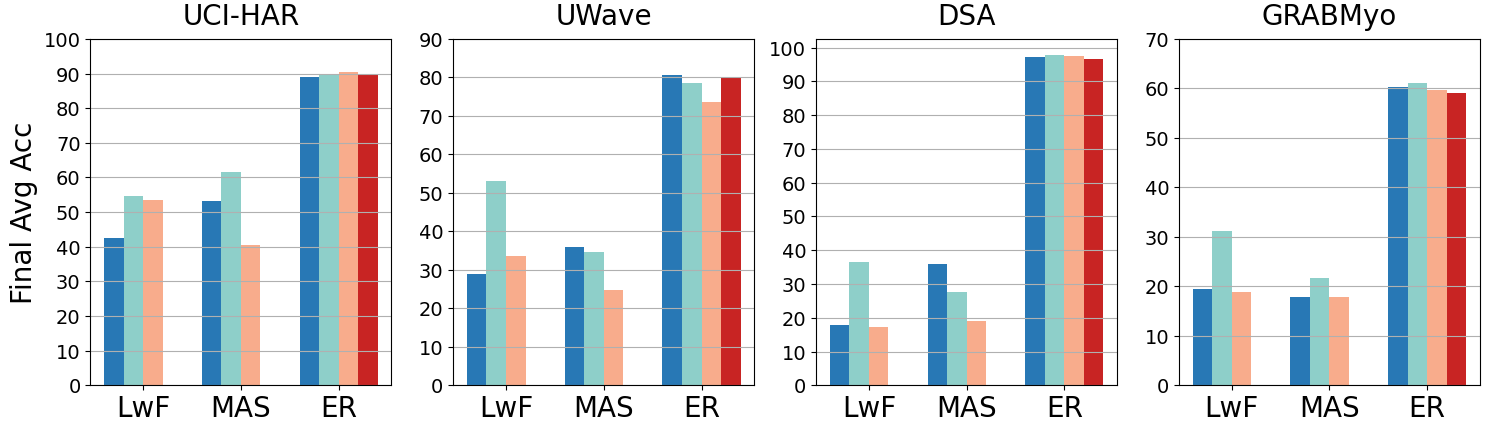}
            \label{figure:abl_cls_ln}
        }
    \end{center}
    \vspace{-6mm}
    \caption{Ablation study on different types of classifiers. The first two represent the single-head classifier trained with CE and BCE, respectively.}
    \label{figure:abla_cls}
\end{figure}

In this section, we investigate the effect of memory budget and classifier type on TSCIL performance. We first evaluate ER-based methods across a range of memory budgets, with results shown in Figure \ref{figure:abl_mem}. The memory budgets are set to 1\%, 5\%, 10\%, 20\% and 100\% of the size of whole training dataset. After that, we compare the performance of LwF, MAS, ER when using 3 different types of classifier. We present the results in Figure \ref{figure:abla_cls}. All the evaluations are conducted using both BN and LN for normalization. The results answer the following questions.

\textbf{Question 3:} \textbf{How does memory budget affect ER-based methods?} As intuitively expected, ER-based methods generally demonstrate increased performance as the memory buffer size grows. However, it's important to note that \textit{the performance gains saturate beyond a certain buffer size}. Such trends exhibit differences between using BN and LN. A surprising observation is at a 100\% memory budget where all encountered data is saved for replay (same as Offline). When using BN, all methods except for ASER show an obvious performance gap compared to offline training. This suggests that the bias of BN towards new tasks does not stem from the imbalance between the memory budget and the size of the new task's training data. Rather, \textit{the bias arises from the 2-batch ER pipeline} (see Algorithm \ref{alg:algo_er} in Appendix \ref{app: alog}). As the number of tasks increases, the proportion of each old class samples in $B_{\mathcal{M}}$ diminishes, leading to an unbalanced distribution of old and new class samples in each step. In contrast, the results on LN show consistent trends in all the configurations, saturating at a level that closely approximates Offline. These outcomes demonstrate a potential issue in the standard ER protocol and further underscore the advantages of using LN for replay in TSCIL.


\textbf{Question 4:} \textbf{How do different classifier types affect TSCIL?} The conventional softmax classifier is known to exhibit a bias problem in CIL scenarios without rehearsal data, that is, the magnitudes of weights of new classes are larger than those of old classes \cite{hou2019learning}. The reason is that minimizing softmax classification loss always reduces the weight magnitudes of the old classes. A simple approach to handle this issue is to replace softmax with sigmoid and train the model with BCE \cite{rebuffi2017icarl, smith2023closer}. Using such a BCE-based classifier, we observe that the results of LwF are consistently improved with a notable margin. However, this improvement is not consistently observed in MAS or ER. Another classifier is Split Cosine Classifier \cite{hou2019learning}, which normalizes the features and class weights and compute their cosine similarity. However, employing such classifier doesn't improve the performance consistently and may even hinder MAS. Lastly, the NCM classifier is suited only for methods utilizing memory samples, and no significant improvement is observed. We posit the reason is that the rehearsal of memory samples mitigates the bias in the single-headed classifier,resulting in a similar performance across different classifiers. In summary, the choice of classifier depends on methods and datasets, and it is less critical for methods employing ER.

\subsection{Evaluation of GR in Privacy-Sensitive Scenarios}
\label{subsec:exp_gr}

In this section, we consider a practical scenario with data privacy concerns, restricting the storage of raw historical samples. As exemplar-free regularization-based approaches exhibit inherent limitations, we investigate GR by using a TimeVAE \cite{desai2021timevae} as the generator. Experiments are run on models utilizing BN or LN, with results presented in Table \ref{tbl:gr_results}.

\textbf{Question 5:} \textbf{How does GR vs ER perform in TSCIL?} In simpler datasets like UCI-HAR and UWave, GR demonstrates substantial efficacy as an alternative to ER. Notably, it consistently outperforms regularization-based approaches and also exhibits comparable or better results to ER. By displaying raw and generated samples from UWave, we find that GR can produce pseudo samples that resemble certain patterns found in the original data. (see Figure \ref{figure:raw_gen_samples}). However, the effectiveness of GR is limited in more complex datasets like DSA and GRABMyo, exhibiting a notable performance gap compared to ER. We attribute GR's limitations to two reasons. First, training a proficient generator model on datasets with large number of classes or variables is still challenging for time series data, especially when the training process is incremental. The diversity of generated samples is also limited (see Figure \ref{figure:more_gr} in Appendix \ref{app:gr_samples}). Secondly, the naive GR method cannot control the class of generated samples, hindering a balanced rehearsal on old classes. In contrast, ER's i.i.d. memory updating circumvents these issues, leading to a pronounced performance differential. Additionally, similar to ER, GR benefits from the use of LN over BN, especially in UCI-HAR and DSA. This indicates that inherent bias impacts all CIL methods employing replay. In summary, while GR shows strong competitiveness in simpler datasets, it encounters notable challenges in more complex environments.

\begin{table}[!ht]
    \centering
    \caption{Evaluation metrics of GR on 4 TSCIL benchmarks.}
    \resizebox{0.98\columnwidth}{!}{
    \begin{NiceTabular}{cccccc}
        \toprule
        \multirow{2}{*}{Norm} & \multirow{2}{*}{Metric}  & \multicolumn{4}{c}{Dataset} \\ \cmidrule{3-6}
         & & UCI-HAR & UWave & DSA & GRABMyo \\
        \midrule
        \multirow{3}{*}{BatchNorm} & $\mathcal{A}_T$ & 62.2{\tiny $\pm$11.1}  & 75.5{\tiny $\pm$8.2} & 37.2{\tiny $\pm$5.9} &  22.4{\tiny $\pm$3.9} \\
                        ~ & $\mathcal{F}_T$ & 45.9{\tiny $\pm$19.6} & 27.8{\tiny $\pm$10.9} & 67.3{\tiny $\pm$11.8} & 69.1{\tiny $\pm$6.6}\\
                        ~ & $\mathcal{A}_{cur}$ & 92.8{\tiny $\pm$10.4} & 96.3{\tiny $\pm$1.6} & 93.2{\tiny $\pm$6.0} & 77.6{\tiny $\pm$4.1}\\

        \midrule

        \multirow{3}{*}{LayerNorm} & $\mathcal{A}_T$ & 81.0{\tiny $\pm$6.7} & 75.8{\tiny $\pm$10.3} & 66.7{\tiny $\pm$14.3} &  20.3{\tiny $\pm$0.7}\\
                        ~ & $\mathcal{F}_T$ & 24.7{\tiny $\pm$14.7} & 29.3{\tiny $\pm$12.9} & 39.9{\tiny $\pm$17.2}  & 94.9{\tiny $\pm$2.8} \\
                        ~ & $\mathcal{A}_{cur}$ & 97.5{\tiny $\pm$3.8} & 97.8{\tiny $\pm$0.8} & 99.9{\tiny $\pm$0.1} & 96.2{\tiny $\pm$1.6} \\

        \bottomrule
    \end{NiceTabular}}
    \label{tbl:gr_results}
\end{table}

\subsection{Analyzing Intra-Class Variations among Subjects}
\label{subsec:exp_sub}

\begin{figure}[t]
    \centering
    \setlength{\abovecaptionskip}{2pt}
    \begin{minipage}[b]{0.38 \columnwidth}
        \centering
        \includegraphics[width=\textwidth]{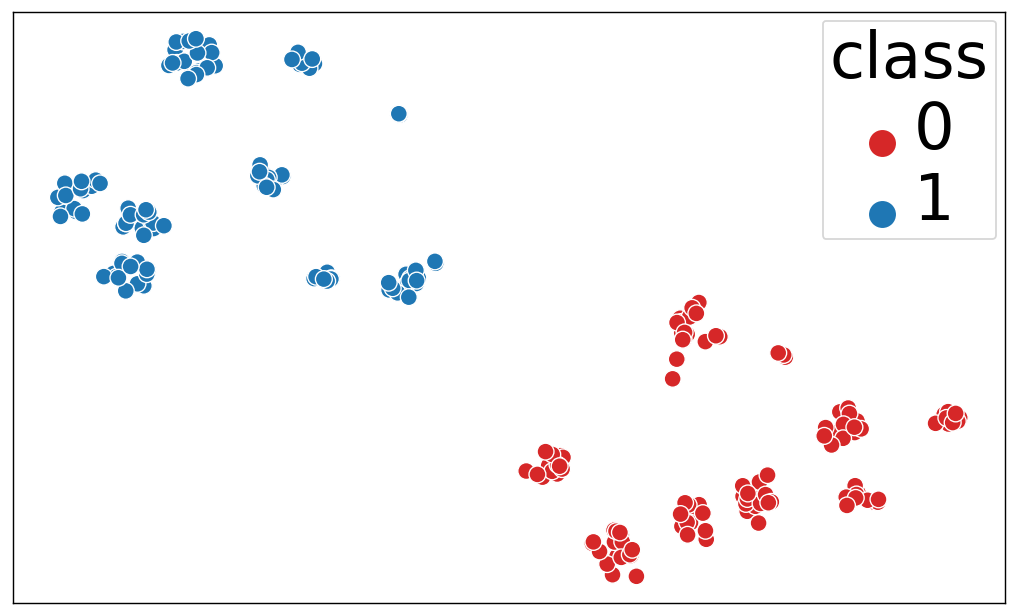}
        \captionof{figure}{Intra-class variations in DSA.}
        \label{fig:intask_dist}
    \end{minipage}%
    \hfill 
    \begin{minipage}[b]{0.58\columnwidth}
        \centering
        \resizebox{\textwidth}{!}{
        \begin{tabular}{cccc}
        \toprule
        Method & $\mathcal{A}_T$ & $\mathcal{F}_T$ & $\mathcal{A}_{cur}$ \\
        \midrule
        ER (2 Sub) & 73.1{\tiny $\pm$25.7} & 24.1{\tiny $\pm$26.1} & 85.8{\tiny $\pm$14.4} \\
        ER (4 Sub) & 76.4{\tiny $\pm$14.4} & 25.8{\tiny $\pm$14.1} & 92.4{\tiny $\pm$6.3} \\
        ER & 79.6{\tiny $\pm$15.6} & 23.6{\tiny $\pm$19.2} & 89.9{\tiny $\pm$9.6} \\
        Balanced & \textbf{87.8}{\tiny $\pm$12.1} & \textbf{13.3}{\tiny $\pm$14.0} & \textbf{93.9}{\tiny $\pm$7.1} \\
        \bottomrule
        \end{tabular}}
        \captionof{table}{Evaluation of subject-based ER variants using BN.}
        \label{tab:intask_results}
    \end{minipage}
\end{figure}

\textbf{Question 6:} \textbf{How do intra-class variations affect TSCIL?} Time series data are commonly collected from various subjects or sources, each may exhibit a distinct input domain. For instance, Figure \ref{fig:intask_dist} depicts the feature distribution within a VAE trained on two classes from DSA. Notably, each class forms eight clusters within the feature space, each corresponding to a different subject. Although such phenomenon is often overlooked in TSCIL, we find that the distribution shift between different subjects may impact the learning performance in a non-trivial level. We further analyse how such subject distribution affects ER-based methods in Appendix \ref{app_subsec_intrav}.

To further investigate this, we use the DSA dataset to compare the original ER baseline with two of its variants. The original ER employs reservoir sampling for \textit{MemoryUpdate}, theoretically ensuring that memory samples in the buffer are i.i.d to the original distribution. However, its \textit{MemoryRetrieval} policy, based on random selection, may not ensure that each batch of replayed samples follows the subject distribution. Our first variant modifies the \textit{MemoryUpdate} policy to select samples only from part of the subjects, deliberately causing the subject distribution of memory samples to deviate from the actual distribution. The second variant maintains the original \textit{MemoryUpdate} policy but improves the \textit{MemoryRetrieval} policy to ensure memory samples are subject-balanced in each retrieved batch. The evaluation metrics are presented in Table \ref{tab:intask_results}, where the first two methods correspond to the first variant sampling from two and four subjects, respectively. "Balanced" represents the second balanced-retrieval variant. 

The observed results underscore the significance of maintaining the subject distribution in TSCIL. Specifically, sampling from part of the subjects demonstrates a diminished replay effect. In contrast, the use of subject-balanced memory samples significantly enhances the rehearsal process. This finding corroborates the idea that incorporating intra-class variations into CIL can improve outcomes. Ignoring this aspect, on the other hand, leads to sub-optimal results. These insights point to an emerging challenge in TSCIL, particularly for methods relying on ER and GR: the need to account for distribution shifts within class caused by different input domains.

\section{Future Directions} \label{sec:future}

This section outlines potential future directions in TSCIL research. 

(1) \textit{Generative Replay for complex time series}: Using GR in complex datasets is a challenge for further exploration. We list several potential solutions. The first is to convert raw time series into time-frequency representations like spectrograms, and to use image-generating models to improve TS synthesis \cite{7952132, 8937236, alaa2021generative}. The second is to apply causality learning, which aims to uncover the underlying data generation processes. Combining it with continual learning emerges as a promising approach to enhance model interpretability and adaptability to distribution shifts \cite{pmlr-v208-churamani23a}, especially when applying it to time series generation \cite{yao2022learning}. The last one is to investigate CIL methods based on \textit{model inversion} \cite{yin2020dreaming,liu2020mnemonics}, which have proven effective in synthesizing pseudo samples in the image domain . 

(2) \textit{Intra-class variations}: The incorporation of intra-class variations into CIL methods are different based on the strategies they use. For regularization, assuming that intra-class variations are similar across different classes \cite{yu2023continual}, a potential avenue is to design a metric to capture the intra-class variations that can be used as a regularization term. For ER, a direction is to tailor memory management policies to take intra-class variations into account. For GR, one may implement an intra-cluster-conditional generator \cite{10204693} to improve the performance.

(3) \textit{Non-standard CIL setup}: This paper focuses on the standard CIL setup. For industry settings, one may need to consider more practical factors beyond our current academic setup, such as data imbalances \cite{kiyasseh2021clinical}, irregular sampling \cite{gupta2021continual}, and online \cite{mai2022online} or mutli-view learning \cite{li2024multi}. We plan to extend our framework to incorporate these challenging settings in the future.

(4) \textit{Incorporation of frequency domain knowledge}: One of critical limitations of current CIL methods lies in the overlook of the intrinsic differences between TS and images. For example, time series are more likely to exhibit periodicity than images. Moreover, TS encapsulates crucial information within the frequency domain or the time-frequency domain. However, existing methods are generic, overlooking these vital properties. Incorporating such properties into TS-specific algorithms is an important topic for future research.

(5) \textit{Time Series Foundation Model}: Large pretrained models has shown competitive performance in image-based CIL \cite{wang2022learning}, even in the absence of memory samples. However, the exploration of pretrained models in TSCIL remains understudied, primarily due to the absence of a universal TS pretrained model. However, the recent advancements in developing Time Series Foundation Models \cite{wu2022timesnet, zhou2023one} mark a significant milestone. Such models are pretrained on a vast collection of TS datasets, which can be applied to various downstream tasks like classification or forecasting. Applying such models for TSCIL is a promising direction to explore.

\section{Conclusion} \label{sec:conclusion}
This paper introduces a unified evaluation framework for Time Series Class-incremental Learning (TSCIL). We provide a holistic comparison to demonstrate the promises and limitations of existing CIL strategies in addressing TSCIL problem. Our extensive experiments evaluate important aspects of TSCIL, including algorithms, normalization layers, memory budget, and the classifier choice. We find that replay-based methods generally demonstrate superiority than regularization techniques, and using LayerNorm instead of BatchNorm significantly alleviates the stability-plasticity dilemma. We further explore some challenges of time series data that are vital to the success of TSCIL. Results and analysis highlight the challenges of normalization, data privacy and intra-class variation, and how they impact the results of TSCIL. We firmly believe our work provides a valuable asset for the TSCIL research and development communities.

\begin{acks}
This research is part of the programme DesCartes and is supported by the National Research Foundation, Prime Minister’s Office, Singapore under its Campus for Research Excellence and Technological Enterprise (CREATE) programme.
\end{acks}

\bibliographystyle{ACM-Reference-Format}
\bibliography{reference}


\appendix

\begin{algorithm}
\footnotesize
\caption{Learning incremental task $t$ for ER-based methods}
\label{alg:algo_er}

\SetCustomAlgoRuledWidth{0.49\textwidth}
\SetAlgorithmName{Algorithm}{}{}
    \SetKwInOut{Input}{Input~}
    \SetKwInOut{Output}{Output}
    \SetKwInput{Initialize}{Initialize}
\Input{Memory $\mathcal{M}$, Parameters $\theta_{t-1}^*$, Training set $\mathcal{D}^t$}
    Transfer model parameters: $\theta_t \leftarrow \theta_{t-1}^*$ \\
    \While {$\text{not converged}$}{
        \For(\tcp*[f]{one epoch}){$B\sim \mathcal{D}^t$}{
        $B_\mathcal{M}\!\!\leftarrow\!$ \retrieval{}($\mathcal{M}$)\\
        $\theta_t\leftarrow~\text{SGD}(B\cup B_\mathcal{M},\theta_t)$ \\
        }
    }
    Obtain optimal parameter $\theta_{t}^*$ \\
    \For(\tcp*[f]{additional pass}){$B\sim \mathcal{D}^t$}{
        $\mathcal{M}\leftarrow$ \update{}$(B, \mathcal{M}, \theta_{t}^*$)\\
        }
    \textbf{return} $\theta_{t}^*$, $\mathcal{M}$
\end{algorithm}

\begin{figure*}[h]
	\begin{center}
		\subfigure[Raw samples]
		{	\includegraphics[width=1.9\columnwidth]{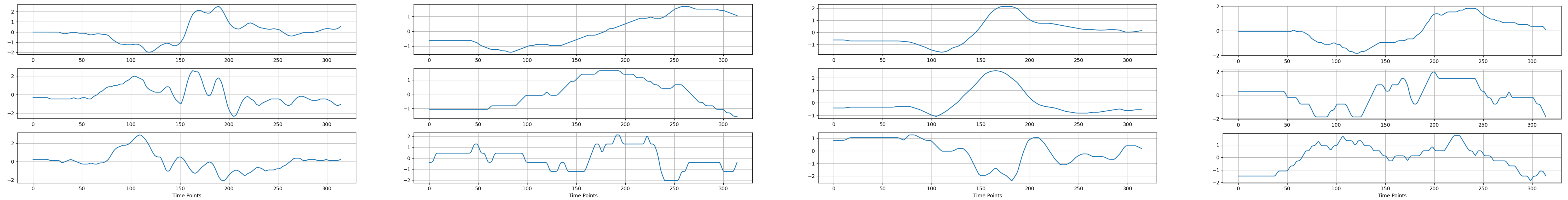}
	   \label{figure:more_raw}	
  }
		\subfigure[Generated samples]
		{	\includegraphics[width=1.9\columnwidth]{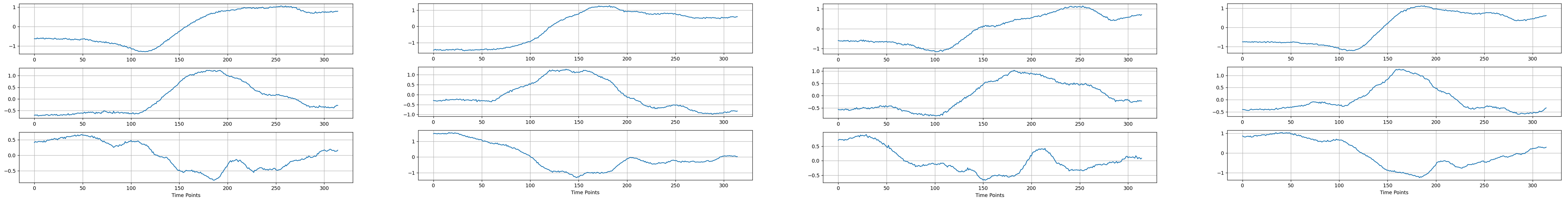}
            \label{figure:more_gen}
		}

	\end{center}
 	\vspace{-5mm}
	\caption{Comparative analysis of raw and GR-generated samples within a single UWave class. While the raw samples in \ref{figure:more_raw} display a diversity of waveforms, the generated samples in \ref{figure:more_gen} tend to exhibit a homogeneous pattern.
	} \label{figure:more_gr}
\end{figure*}

\section{Details of the framework}
\label{app:framework_details}
\subsection{Details of algorithms}
\label{app: alog}
Here we introduce some details of ER, GR and TS-specific methods. Other generic methods follow their formal implementations \cite{JMLR:v24:23-0130}. (1) \textbf{ER}: The ER-based learning pipeline is outlined in Algorithm \ref{alg:algo_er}. Baisc ER ~\cite{rolnick2019experience, chaudhry2019continual} uses Reservoir Sampling ~\cite{vitter1985random} as \textit{MemoryUpdate} policy and random selection as \textit{MemoryRetrieval} policy.
(2) \textbf{GR}: We follow \cite{shin2017,van2022three} 
 to train a generator $g_{\phi}$ along with the learner $f_{\theta}$. Before task $t$, copies of $f_{\theta_{t-1}^*}$ and $g_{\phi_{t-1}^*}$ are saved. In each step, a batch of pseudo samples are generated and assigned labels by $f_{\theta_{t-1}^*}$. We train $g_{\phi_{t}}$ after training of $f_{\theta_{t}}$ is complete. Once the task is over, the saved copies are updated. 
(3) \textbf{DT2W}: DT$^2$W \cite{10094960} employs Soft-DTW \cite{cuturi2017soft} to distill intermediate feature maps. In addition to this temporal KD loss, it integrates LwF to regularize the whole model. To calibrate the bias in the classification head, the method incorporates feature replay via Prototype Augmentation \cite{Zhu_2021_CVPRb, 10.1145/3581783.3611926}.
(4) \textbf{CLOPS}: Original CLOPS \cite{kiyasseh2021clinical} tailors both \textit{MemoryUpdate} and \textit{MemoryRetrieval} policies. However, we observe that their \textit{MemoryRetrieval} policy encounters a limitation as replayed samples are updated after multiple epochs, potentially leading to overfitting and diminishing the effectiveness of replay. In response to this, we opt to employ random selection for memory retrieval in our experiments. 
(5) \textbf{FastICARL}: It is a fast variant of iCaRL, replacing herding with a \textit{MemoryUpdate} strategy based on KNN and a maxheap. We omit the quantization process which compresses memory samples for a fair comparison.

\subsection{More implementation details}
\label{app: imple details}
 For early stopping, we set a patience of 20 for ER-based and GR-based methods, and a patience of 5 for other methods. During the learning of each task, a separate validation set is split from the training data of that task, with a split ratio 1:9. The earlystopping is triggered based on the validation loss calculated on that split validation set. For hyperparameters, we tuned the learning rate and batchsize and consistently found that the best configuration was 0.001 and 64. Following \cite{nie2022time}, the choice of the learning rate scheduler is set as an hyperparameter, selecting from the following strategies: (1) step10, (2) step15 and (3) OneCycleLR \cite{smith2019super}. The first two scheduler decay the learning rate by 0.1 after 10 epochs and 15 epochs, respectively. For UCI-HAR and UWave, we follow \cite{10094960} to set the dropout rate to 0, while for other datasets, it is set at 0.3. For methods with replayed samples, we set the batchsize to 32 for both $B$ and $B_\mathcal{M}$, ensuring the total number of samples used per step is the same across all the methods. For each run, we tune the best hyperparameters by running 2 validation runs on validation tasks. Detailed hyperparameter grids can be found in the code base. 

\section{Further Experiment Results}
\label{app:sec more results}

\subsection{Visualization of synthesized samples of GR}
\label{app:gr_samples}

\begin{figure}[h]
    \begin{center}
        \subfigure[Raw samples]
        {
            \includegraphics[width=.96\columnwidth]{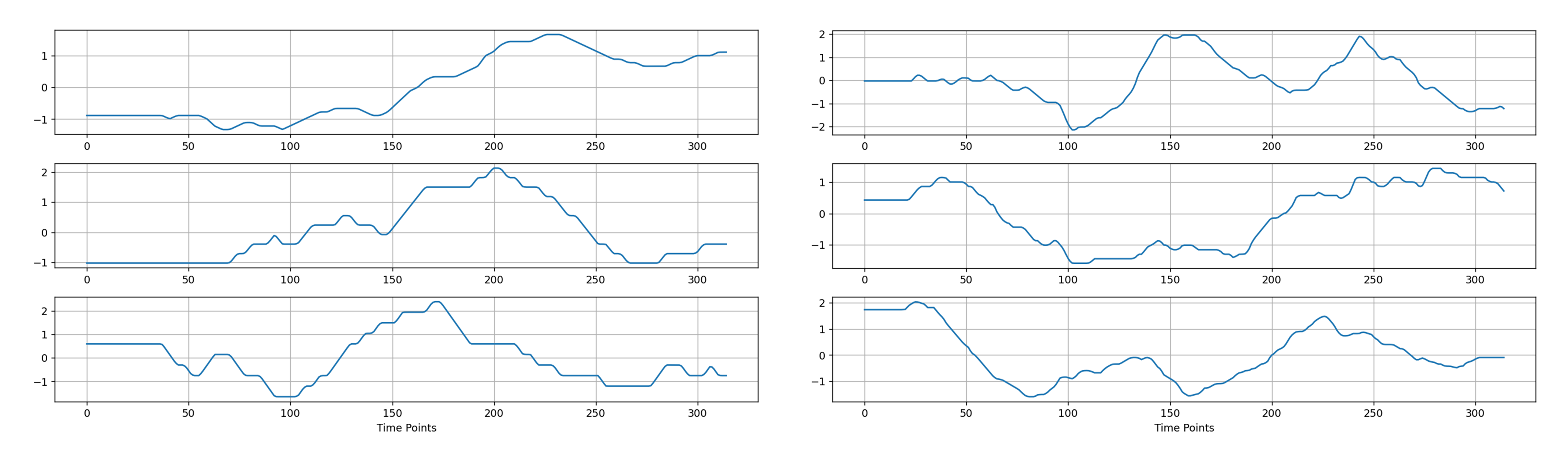}
            \label{figure:raw_samples}
        }
    \end{center}
    \begin{center}
        \subfigure[Generated samples]
        {
            \includegraphics[width=.96\columnwidth]{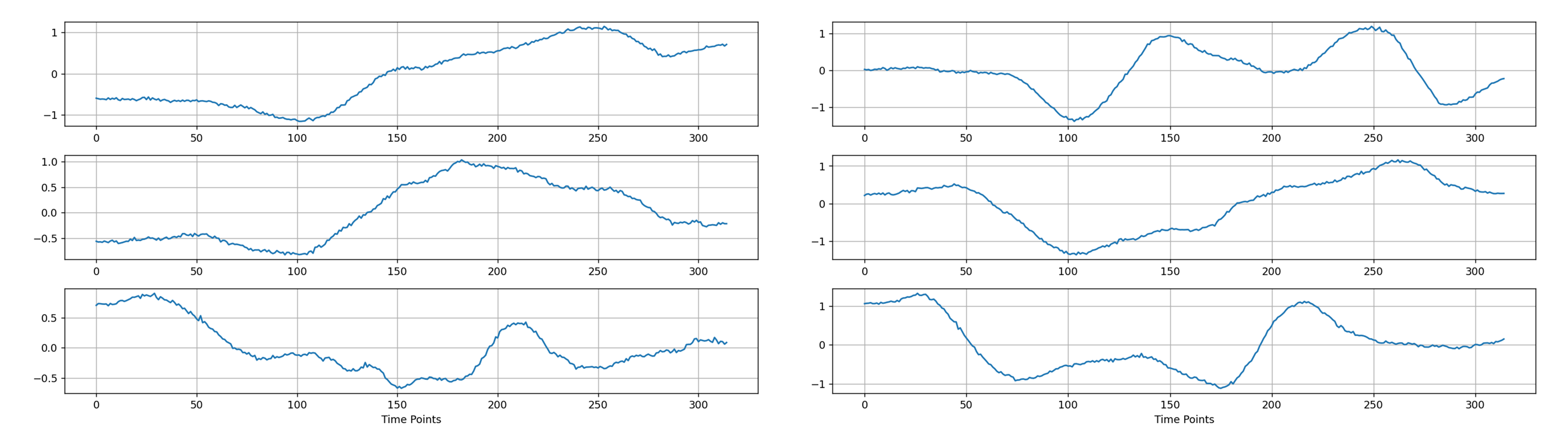}
            \label{figure:gen_samples}
        }
    \end{center}
    \vspace{-5mm}
    \caption{Comparison of raw and GR-synthesized UWave samples across 2 classes. Left and right columns represent distinct classes respectively.}
    \vspace{-5mm}
    \label{figure:raw_gen_samples}
\end{figure}

To reflect the effect of generation, we show some raw and generated samples from the UWave dataset. We select this dataset since the its samples only have 3 variables, which are easy to visualize and compare. Figure \ref{figure:raw_gen_samples} displays examples of raw and generated samples from two UWave classes. Specifically, the generator model is the optimized version right after learning the first task, i.e. $g_{\phi_{1}^*}$. The generator is used to synthesize the two classes in the first task. We note that the generated samples are unlabelled and their class labels are determined by passing them to the classification model $f_{\theta_{1}^*}$. We can see that the generated samples are similar to the shown original ones, demonstrating that GR is capable of synthesizing some real patterns in simple TS data.

Apart from the above, we also investigate the diversity of the generated samples. Similarly, we focus on the UWave dataset and use the generator model after learning the first task. Unlike previous comparisons, we only examine the raw and generated samples from a single class. Figure \ref{figure:more_gr} selectively depicts 8 samples from the single UWave class, comprising 4 raw and 4 generated samples. We can observe that raw samples appear to exhibit several different patterns, demonstrating the diversity present in the original data. In contrast, generated sample prone to exhibit a few similar patterns. This indicates that a potential challenge for using GR in TSCIL is to ensure the diversity of generated data.

\subsection{How do intra-class variations affect TSCIL?}
\label{app_subsec_intrav}
To further understand the role of intra-class variations caused by subjects, we consider a simple scenario that using ER to learn a sequence of 2 tasks. Here we assume that both tasks are sourced from the same group subjects $S=\{s_1, \dots, s_k\}$, with a prior distribution as  $p(S) = \sum_{s \in S}p(S = s)$. The input distribution is conditioned on subject $s$ and can be represented as $p(X) = \sum_{s \in S} p(X | S = s) p(S = s)$. For incrementally learning the two tasks with distributions $p(\mathcal{X}_1, \mathcal{Y}_1)$ and $p(\mathcal{X}_2, \mathcal{Y}_2)$, the learning objective can be formulated as:

\begin{equation}
\small
\begin{aligned}
    \max & \sum_{s \in S} p_{1}(s) \sum_{(\mathbf{x}_1, y_1) \in \mathcal{D}_1} \log p(y_1 | \mathbf{x}_1, s) \\
     + & \sum_{s \in S} p_{2}(s) \sum_{(\mathbf{x}_2, y_2) \in \mathcal{D}_2} \log p(y_2 | \mathbf{x}_2, s)
\end{aligned}
\label{subject_obj}
\end{equation}
\noindent where $p_{1}(s)$ and $p_{2}(s)$ represent the distribution of subjects in task 1 and task 2, respectively. When learning the second task, the second term is calculated on incoming training set thus it is clear that  $p_{2}(s)= p(s)$. However, the first term is computed based on replaying memory samples. But typical memory management strategies tend to overlook the implicit influence of subject $s$ and omit to maintain $p_{1}(s)= p(s)$, making the replayed data deviated from the original one. This problem can be extended to general cases with more tasks, showing the influence of variations caused by different subjects.

\end{document}